\pdfoutput=1

\documentclass[11pt]{article}

\usepackage[final]{acl}

\usepackage{times}
\usepackage{latexsym}

\usepackage{booktabs} 
\usepackage{kotex} 
\usepackage{tabularx} 
\usepackage{multirow} 
\usepackage{xcolor}
\definecolor{summer3}{RGB}{164,212,149}  
\definecolor{summer1}{RGB}{214,235,157}  
\definecolor{summer2}{RGB}{212,212,212}     
\usepackage{amssymb}
\usepackage{ragged2e} 

\usepackage[T1]{fontenc}

\usepackage[utf8]{inputenc}

\usepackage{microtype}

\usepackage{inconsolata}

\usepackage{graphicx}

\title{KpopMT: Translation Dataset with Terminology for Kpop Fandom}

\author{JiWoo Kim \\
  Sungkyunkwan University \\
  Suwon, South Korea \\
  \texttt{wldn9705@skku.edu} \\\And
  Yunsu Kim \\
  aiXplain Inc. \\
  Los Gatos, USA \\
  \texttt{yunsu.kim@aixplain.com} \\\And
  JinYeong Bak \\
  Sungkyunkwan University \\
  Suwon, South Korea \\
  \texttt{jy.bak@skku.edu}}

\begin{document}
\maketitle
\begin{abstract}

While machines learn from existing corpora, humans have the unique capability to establish and accept new language systems. This makes human form unique language systems within social groups. Aligning with this, we focus on a gap remaining in addressing translation challenges within social groups, where in-group members utilize unique terminologies. We propose KpopMT dataset, which aims to fill this gap by enabling precise terminology translation, choosing Kpop fandom as an initiative for social groups given its global popularity. Expert translators provide 1k English translations for Korean posts and comments, each annotated with specific terminology within social groups' language systems. We evaluate existing translation systems including GPT models on KpopMT to identify their failure cases. Results show overall low scores, underscoring the challenges of reflecting group-specific terminologies and styles in translation. We make KpopMT publicly available.\footnote{\url{https://github.com/skswldndi/KpopMT}}

\end{abstract}

\section{Introduction}
\label{sec:introduction}
One of the most profound distinctions between humans and machines lies in their ability to form a new language system. While machines learn from and rely on existing corpora, humans have the unique capability to establish new social conventions, such as agreeing to call an apple ``apple''. This inventive ability is deeply intertwined with the social nature of language, wherein words gain influence and become integrated into the vernacular of social groups -- termed as social dialect -- through social interactions and contexts. 

In light of this, our research places a particular emphasis on specific phenomena: Social groups often develop their unique linguistic systems, replete with specific terminologies and jargon, as in Table \ref{tab:socialgroup-example} \citep{wolfram2004social,peterson2014linguistics}. The global surge in social media usage has brought individuals from diverse countries and continents together, forming cohesive communities \citep{sawyer2012impact}. An example is the Kpop fandom across various regions \citep{choi2014south}; these fans seek to connect and establish interpersonal relationships that transcend language barriers \citep{malik2021english}. 


\begin{table}[t]
    \footnotesize
    \centering
    \begin{tabularx}{\columnwidth}{lX} 
    \toprule
    \multicolumn{2}{c}{Kpop Fandom Group's Language System} \\
    \midrule
    \multirow{2}{*}{Korean}  & 원래 \colorbox{summer3}{덕질}할 때 \colorbox{summer2}{갠팬} 수준으로 \colorbox{summer1}{최애} 위주로만 파는 타입이었음 \\
    \midrule
    \multirow{3}{*}{English} & Originally, when I \colorbox{summer3}{stan}, I tended to focus on only \colorbox{summer1}{my bias} almost it seems like \colorbox{summer2}{solo stan}. \\
    \bottomrule 
    \end{tabularx}
\caption{Example of how specialized terminologies (colorboxes) are used in a global social group. }
\label{tab:socialgroup-example}
\end{table}

However, the unique linguistic systems of social groups are currently not adequately represented in machine translation (MT) systems. Despite proposals from researchers to consider language diversity in MT \citep{kumar2021machine, lakew2018neural}, including cross-domain standard languages \citep{hu2019domain, currey2017copied}, local dialects \citep{abe2018multi,hassan2017synthetic}, and related languages \citep{pourdamghani2017deciphering}, the specialized language systems of social groups remain underexplored in MT research. For example, Google Translator translates the example sentence in Table \ref{tab:socialgroup-example} as \textit{Originally, when I was a fan, I focused on my favorites. It was the type that only sells.}, which differs from the specialized language system of the group. 

In order to promote research in this direction, we argue for the necessity of a benchmark dataset that encompasses the language systems of social groups. This dataset is also crucial for demonstrating the capabilities of the current translation systems specifically tailored to handle the nuances of social group languages.

In our study, we reference the field of terminology-based MT \citep{xie2020constraining, knowles2023terminology}, which focuses on accurately translating specialized terminology in machine-generated output. Notably, the medical domain has seen significant advancements in this area \citep{alam2021findings, anastasopoulos2020tico}, highlighted by the release of datasets that aid the NLP community, particularly those interested in terminology translation and constrained neural MT tasks \citep{zhang-etal-2023-disambiguated, wang2022template}.

We propose a terminology-tagged MT dataset tailored for social groups, named \textbf{KpopMT}. Through human survey, we demonstrate participants in a social group show a strong preference for reference translations in KpopMT compared to translations without terminology. We choose the Kpop fandom as the social group for this initiative, given its global popularity and the widespread sharing of content on social media platforms among fans from diverse countries \citep{ringland2022army}.

KpopMT is collected from Korean posts and comments found on fan-related websites, including X (Twitter). Expert translators fluent in the terminologies used within this social group provide English translations, resulting in 1k sentence pairs, each annotated with specific terminologies. 
We evaluate existing translation systems, including state-of-the-art models such as GPTs, on KpopMT to identify areas of improvement and establish baseline performance for future research. 

\begin{table*}[t]
    \centering
    \small
    \begin{tabularx}{\textwidth}{p{0.10\textwidth}X}
        \toprule
        \multicolumn{2}{l}{1. Sentence Phase} \\
        \toprule
        \multirow{2}{*}{Source} & \colorbox{summer1}{트친}이랑 푸드코트에서 밥 먹다가 \colorbox{summer2}{용수} 만났어요 \colorbox{summer3}{포카} 보여주고 팬이라고 싸인도 받았어요 덕계못인줄 알았는데\\
        \midrule
        \multirow{2}{*}{Reference}    & I met \colorbox{summer2}{Yongsoo} at food court while eating with my \colorbox{summer1}{moot}. We showed him \colorbox{summer3}{pc} and got an autograph saying that we're fans of him. I thought \textit{\underline{stan}} can't get luck to see \textit{\underline{faves}} \\
        \toprule
        \multicolumn{2}{l}{2. Terminology Phase} \\
        \midrule
        Tag 1
        & <term id="202" type="slang" source="\colorbox{summer1}{트친}" target="\colorbox{summer1}{twitter moots|moots|moot}"> \colorbox{summer1}{moot} </term> \\
        \midrule
        Tag 2  & <term id="146" type="group-NE" source="\colorbox{summer2}{용수}" target="\colorbox{summer2}{Yongsoo}"> \colorbox{summer2}{Yongsoo} </term> \\
        \midrule
        Tag 3  & <term id="8" type="group-lexicon" source="\colorbox{summer3}{포카}" target="\colorbox{summer3}{pocas|poca|pcs|pc}"> \colorbox{summer3}{pc} </term> \\
        \bottomrule
    \end{tabularx}
    \caption{Example of KpopMT. }
    \label{tab:annotationexample}
\end{table*}


\section{KpopMT}
\label{sec:dataset}
KpopMT consists of three parts: 1) the parallel sentence tagged with terminologies, 2) termbase which contains parallel glossary, and 3) fandom monolingual dataset. In this section, we illustrate how the parallel sentence and termbase are constructed. We address fandom monolingual dataset in Section \ref{subsec:experimentsetting}.

In Table \ref{tab:annotationexample}, we show an example of KpopMT. Capturing the most distinctive part of social groups' language system is terminology, we also provide terminology information. The parallel sentence has fandom-related terms included in both the source and target side, which are annotated as tags. 

To ensure translation reliability, we get confirmation from five native English Kpop fans who know both Korean fandom terms and English fandom terms. 

\subsection{Construction}
\label{subsec:collection}
KpopMT is constructed in two phases. First, we construct parallel sentences. Second, we annotate parallel terminology information in the sentences. 

\paragraph{Sentence Phase}
To obtain sentences that include fandom-related terminology, we make a query list derived from crowd-sourced dictionaries.\footnote{Naver Open-Dict \url{http://naver.me/xeUyZv8N}} Using the query list, we manually collect Korean monolingual data from the fan community sites\footnote{Theqoo \url{http://theqoo.net}}\footnote{Instiz \url{http://instiz.net}} and Twitter. Then we hire ten human translators who pass the qualification test, by asking for English translations of ten Korean fandom terms and their meanings. We only hire translators who answer correctly for at least eight terms. Then we ask them to translate Korean sentences into English sentences, resulting in 1,000 sentence pairs. 

Please note that we ask them to translate with the inclusion of English fandom terms, such as ‘stan.' If there are no specific English fandom terms, we ask the translators to include internet slang in the sentence as a variation of standard language. We also forbid the usage of translation services such as Google Translate. 

\paragraph{Terminology Phase}

In this phase, we mark terms included in the parallel sentences first. After extracting all marked terms, we create a parallel glossary by matching Korean terms with their English counterparts. If there are other possible or morphological variations in English terms, we include them on the target of the glossary, separating them with ‘|'. Subsequently, experts on the terminology used within fandom and internet slang confirm the consistency of parallel glossary. 

Since we separate the sentence phase and the terminology phase, KpopMT is not structured according to the terminologies in this glossary. This complicates the creation of a one-to-one dictionary. For instance, `머글' is not a direct equivalent of only `local'. Therefore, we opt to begin with the Korean side as it aligns with the actual translation direction in which the data was generated, following \citet{alam2021findings}. 

We categorize the glossary terms into three: \textbf{Group-Lexicon}, \textbf{Group-NE}, and \textbf{Slang}. Group-Lexicon refers to the lexicon unique to the fandom, which may not be understood by those outside the fandom. Group-NE represents elements related to the fandom's named entities, such as idol group names or nicknames. Slang encompasses internet slang, which is a variation of standard language. We only apply truecasing to Group-Lexicon and not to Group-NE and Slang, as capitalization is crucial for names and we believe there may be different meanings in truecased slang (e.g., ‘ig' and ‘IG'). 

In the end, we tag the sentence pairs on both source and target side with possible terminology translations from the glossary. We tag the sentence pairs with terminology translations only if both source terminology and a corresponding target terminology exist in the reference translation, following \citet{alam2021findings}. So, we do not tag ‘stan' and ‘faves' terms in Table \ref{tab:annotationexample}.

\subsection{Key Characteristics}
\label{subsec:data-analysis}
The parallel corpus of KpopMT contains 6k Korean tokens and 12k English tokens, excluding the tags. It has a substantial number of tagged sentences, totaling 1,035, among which the tags categorized as Group-Lexicon are the most prevalent, accounting for 858 (82.8\%), followed by Group-NE with 92 (8.8\%), and Slang with 85 (8.4\%). This indicates that KpopMT contains substantial information on a specific social group, which is Kpop fandom.

In addition, to assess its suitability for evaluating terminology-based translation, we compare KpopMT with TICO-19, widely recognized as the most common terminology machine translation dataset (Table \ref{tab:comparewithtico}) \cite{anastasopoulos2020tico, odermatt2023cascaded}. It is important to note that significant portions of TICO-19 lack any terminological content. KpopMT encompasses more extensive meaningful portions of terms compared to TICO-19. 

\begin{table}[t]
    \centering
    \small
    \begin{tabularx}{\columnwidth}{lrrrrr}
    \toprule
            & \multicolumn{5}{c}{Number of Terms}                                                                                                 \\
    Dataset & \multicolumn{1}{c}{0}                          & \multicolumn{1}{c}{1} & \multicolumn{1}{c}{2} & \multicolumn{1}{c}{\textgreater{}=3} & \multicolumn{1}{c}{max} \\
    \midrule
    Ours    & \multicolumn{1}{r}{27.5\%}      & \textbf{47.7}\%                     & \textbf{19.8}\%                     & \textbf{5.0}\%                                    & 5                       \\
    TICO-19 & \multicolumn{1}{r}{\textbf{40.2\%}} & 22.6\%                & 6.6\%                 & 3.9\%                                & 9                      \\
    \bottomrule
    \end{tabularx}
\caption{Comparision with TICO-19 regarding number of terminologies in each line. } 
\label{tab:comparewithtico}
\end{table}

Moreover, to demonstrate the necessity of KpopMT, we conduct a human survey involving five native English-speaking Kpop fans. None of them are involved in data construction process. 
They are given with 80 English sentences, each of which has two versions: samples from KpopMT with fandom-specific terms and expressed in standard language. Fans strongly prefer translations with fandom terms (89.75\%) when asked which ones make them feel more connected to fandom members. This highlights the importance of considering cultural factors in communication \citep{gudykunst2003cross}, making KpopMT valuable for both translation accuracy and social connectedness.



\section{Experiments}
\label{sec:experiments}
We evaluate existing machine translation systems on KpopMT to assess its difficulty. 

\subsection{Experimental Setting}
\label{subsec:experimentsetting}
\paragraph{Data}
In our experiments, we utilize two types of data: general standard language data and fandom language data. To acquire the general standard language data, we download a Korean-English dataset (800k) from AI Hub, which is a platform releasing AI data by the Korean government.\footnote{\url{https://www.aihub.or.kr}} 
To obtain the fandom language data, we scrape 40k monolingual data samples for each language from fan-related websites, employing the query list specified in \ref{subsec:collection}. Korean monolingual data has 287k tokens with an average sentence length of 29.79, while English monolingual data has 376k tokens with an average length of 49.61. For evaluation purposes, we employ a test split comprising 500 sentences from our parallel KpopMT.

\paragraph{Baselines} 
We implement two kinds of baselines: open-source machine translation models and proprietary machine translation systems. Our baseline choice considers three factors: general standard language data trained with, fandom monolingual data trained with (Table \ref{tab:opensourcebaselines}), and state-of-the-art systems. 

\begin{table}[]
    \centering
    \resizebox{\linewidth}{!}
    {
    \begin{tabular}
    {cccc}
    \toprule
    \multirow{2}{*}{Systems} & Fandom & General & Finetuned \\
        & Data & Data & from \\
    \midrule
    \multicolumn{1}{l}{M2M} & & \checkmark & \\
    \midrule
    \multicolumn{1}{l}{mBART w/o Fandom} & & \checkmark & mBART \\
    \midrule
    \multicolumn{1}{l}{mBART w/ Fandom} & \checkmark & \checkmark & mBART \\
    \midrule
    \multicolumn{1}{l}{SL-MT
    } & & \checkmark & \\
    \midrule
    \multicolumn{1}{l}{Domain Adaptation} & \checkmark & \checkmark & SL-MT \\
    \bottomrule
    \end{tabular}
    }
\caption{Features of open-source baselines.}
\label{tab:opensourcebaselines}
\end{table}



\begin{table*}[t!]
    \centering
    \small
    \begin{tabular}{cl|rr|rrr}
    \toprule
                                 &                                                                                       & \multicolumn{2}{c}{Terminology-focused}                                     & \multicolumn{3}{|c}{Translation Quality}                                           \\
                                 & Systems                                                                                & \multicolumn{1}{c}{EMA} & \multicolumn{1}{c}{1-TERm} & \multicolumn{1}{|c}{BLEU} & \multicolumn{1}{c}{COMET} & \multicolumn{1}{c}{chrF++} \\
    \midrule
    \multirow{5}{*}{Open Source} & M2M                                                                                   & 2.4                                     & 10.5                            & 4.1                      & 51.2                     & 19.9                       \\
                                 & mBART w/o Fandom                                                                       & 6.7                                     & \textbf{13.7}                            & \textbf{10.4}                     & \textbf{60.5}                     & \textbf{32.3}                       \\
                                 & mBART w/ Fandom                                                                    & 6.4                                     & 12.3                            & 9.1                      & 56.6                     & 29.0                       \\
                                 & SL-MT                                                                  & 3.8                                     & 6.5                              & 8.7                      & 56.4                     & 30.3                       \\
                                 & Domain Adaptation                                                                     & \textbf{13.9}                                    & 13.3                             & 9.7                      & 57.7                     & 31.2                       \\
    \midrule
    \multirow{3}{*}{Proprietary} & GPT-3.5-turbo-0613                                                                     & 19.3                                    & 1.1                             & 8.7                      & 65.8                     & 32.9                       \\
                                 & GPT-4-0613                                                                           & \textbf{26.4}                                    & 10.3                            & 9.9                      & \textbf{65.8}                     & 35.0                       \\
                                 & \begin{tabular}[c]{@{}l@{}}Google Translator \end{tabular} & 10.4                                    & \textbf{16.3}                            & \textbf{13.9}                     & 65.4                     & \textbf{35.9}                      \\
    \bottomrule
    \end{tabular}
    \caption{Result on test split set using existing machine translation systems. }
\label{tab:result}
\end{table*}

Regarding general standard language data, we set baselines of M2M, mBART w/o Fandom and Standard Language MT (SL-MT). M2M is a multilingual translation model that can translate between any pair from 100 languages \citep{fan2021beyond}. mBART w/o Fandom is a finetuned translation model from multilingual Bart (mBART) using general data \citep{liu2020multilingual}. mBART is the state-of-the-art model on IWSLT-17 \citep{cettolo-etal-2017-overview, park2021should}. SL-MT is a Korean-English state-of-the-art translation model on general standard language data \citep{park2020empirical}.

Regarding fandom monolingual data, we set baselines of mBART w/ Fandom and Domain Adaptation. mBART w/ Fandom is a finetuned translation model from mBART with an injection of both general parallel data and fandom monolingual data. We use the back-translation technique to make pseudo-parallel data for fandom monolingual data \citep{sennrich2015improving}. Domain Adaptation is a finetuned translation model from Standard Language MT, with the technique of iterative back-translation of fandom monolingual data \citep{hu2019domain, dou2019unsupervised}. The differences from mBART w/ Fandom are the pretrained model and iteration of back-translation. In a preliminary study, we find that mBART w/ Fandom's performance drops with iterative back-translation. 

For proprietary machine translation systems, we experiment with OpenAI's GPT models (GPT-3.5 and GPT-4) and Google Translator. For OpenAI's GPT models \citep{openai2023gpt4, eloundou2023gpts}, we assign the role of a Kpop fan to the model. This fan is familiar with terminologies used within the Kpop fandom and internet slang, which has shown empirically the best performance. We provide the prompt in Korean.

\paragraph{Evaluation} 
We evaluate systems on both translation accuracy and terminological accuracy. Translation accuracy is evaluated with standard reference-based MT metrics: SacreBLEU \citep{post2018call, papineni2002bleu}, COMET \citep{rei2020comet}, chrF++ \citep{popovic2017chrf++}. Terminology accuracy is evaluated with exact-match term accuracy (EMA) and 1-TERm score \citep{anastasopoulos2021evaluation}. EMA is an accuracy score that searches for exact term translation matches (of the terminology required output) over the original hypothesis. The 1-TERm score is a modification of the TER metric, biased to assign higher edit cost weights for words belonging to a term (and then simply reversed so that a higher score is better). When computing terminology targeted evaluation, we consider synonyms which are split by ‘|' in the tags. We rank systems according to EMA.

\subsection{Results}
\label{subsec:results}
Table \ref{tab:result} shows overall low scores, underscoring the considerable challenges posed by KpopMT in terms of both terminology-focused content and translation quality.

\paragraph{GPTs}

Overall, the GPT models demonstrate higher scores than other systems. When examining the words they succeed in generating, they excel in producing certain words compared to other models. ‘bias,' ‘ult,' and ‘stan' are frequently generated, contributing to GPTs' higher EMA scores.

However, Group-Lexicon type faces challenges, especially when excluding those basic words.

We conduct an analysis to determine which types of terms GPT-4 could generate or not. 
We calculate each type's EMA score, resulting in Group-Lexicon 16.3\%, Group-NE 21.54\%, and Slang 20\%. As seen in overall low scores in success translation rate for all types, GPT-4 struggles with generating terminologies, particularly those related to Group-Lexicon. Given the importance of 'Group-Lexicon' in Kpop social groups, this deficiency shows a clear need for better translation in specific social contexts. 

\paragraph{General Data}
We ascertain whether a translation model trained on general standard language data could grasp specific language systems within social groups. Although mBART w/o Fandom displays better performance in translation quality evaluations, its EMA score remains low, indicating a lack of specific knowledge pertaining to social groups.

\paragraph{Fandom Data}
Our findings suggest that incorporating fandom data into the training process does not consistently yield improved results. 
The possible explanation is fandom monolingual data during the back-translation process is noisy, resulting in pseudo-parallel data that significantly deviated from general data, which seems to have hindered proper comprehension. Following \citet{michel2018mtnt} and calculating the perplexity score of fandom monolingual data using a language model trained with general data, we get 1395.4 for Korean and 713.8 for English. 
This limitation might prevent effective inference of the meaning and context of relevant terminology.

\paragraph{Translation Quality}
As highlighted by previous research \cite{pascual2020directed}, there exists a tension between fluency of the translation and terminology accuracy in our results. 
Based on this, we intend to explore in future work whether it is possible to enhance the overall translation quality while preserving social groups' distinctive terminology.

\section{Conclusions}
\label{sec:conclusion}
Our approach acknowledges the dynamic and evolving nature of language, especially in digitally mediated communities, which are often underrepresented in traditional linguistic resources. 
We propose a benchmark dataset for social groups' language systems, named KpopMT. 1k parallel dataset contains not only Korean-English parallel sentences but also terminology information of Kpop fandom group. We also make termbase and monolingual data publicly available. We evaluate existing translation approaches on KpopMT to identify their failure cases. Our future plan includes expanding KpopMT to encompass other social groups, such as sports and global movie communities.


\section*{Limitations}
\label{sec:limitations}
In the study, we view Kpop fandom as a broad spectrum and do not focus on a specific fandom of any one idol group. There are numerous Kpop Idols, such as BTS, Twice, Seventeen, etc. and each terminology of their fans is distinct from one another. For instance, BTS fans use \textit{Borahae} to mean \textit{I love you}, while Seventeen fans use \textit{horanghae} to mean the same thing. 

\section*{Ethical Considerations}
\label{sec:ethics}
The created dataset will be released under the terms of the Twitter API. Aside from Tweets, which are sourced from publicly accessible websites, we ensure that there is no violation of copyright or invasion of privacy. To prevent user tracking, we remove any information related to users. All of our dataset is made publicly available through a Creative Commons CC BY-SA 4.0 license.

We compensate volunteer translators with more than the minimum wage in Korea. They are fully aware of their tasks, and we provide them with detailed annotation instructions. After compensation, we anonymize and remove their personal information. Additionally, we carefully filter out data containing hate speech about celebrities to ensure that human translators do not face any risks or harm associated with their participation.

\section*{Acknowledgments}
We would like to thank the anonymous reviewers for their helpful questions and comments. We further would like to express our gratitude to Jonas Belouadi for discussions, proofreading, and comments on our work.  
This research was partly supported by the Bio \& Medical Technology Development Program of the National Research Foundation (NRF) funded by the Korean government (MSIT) (NRF-2021M3A9E4080780, RS-2023-00252083),  
Institute of Information \& communications Technology Planning \& Evaluation (IITP) grant funded by the Korea government (MSIT) (No.2019-0-00421, AI Graduate School Support Program (Sungkyunkwan University).
\bibliography{custom}

\appendix



\end{document}